\documentclass[runningheads]{llncs}
\usepackage{graphicx}
\usepackage{pgf, tikz} 
\usetikzlibrary{arrows, automata} 

\usepackage{amsthm} 
\usepackage{amsmath} 
\usepackage{amssymb} 
\usepackage{enumitem} 

\usepackage[normalem]{ulem}  

\usepackage{xcolor}  
\usepackage[linesnumbered,ruled,vlined]{algorithm2e} 

\SetCommentSty{mycommfont}  

\SetKwInput{KwInput}{Input}  
\SetKwInput{KwOutput}{Output}  

\newcommand{\stkout}[1]{\ifmmode\text{\sout{\ensuremath{#1}}}\else\sout{#1}\fi}

\newcommand{\ninst}{\ensuremath{\dot{\ell}}} 
\newcommand{\vepsilon}{\ensuremath{\log_2|\Upsilon|}}

\usepackage{hyperref}

\newcommand{\sidenoteJ}[1]{}

\newcommand{\sidenoteM}[1]{}

\newcommand{\sidenoteExtra}[1]{}

\usepackage[misc]{ifsym}

\begin{document}
\title{Conditional Teaching Size}
\toctitle{Conditional Teaching Size}
%
%
%
\author{Manuel Garcia-Piqueras (\Letter)\inst{1}\orcidID{0000-0001-8088-8393} \and
Jos{\'e}~Hern{\'a}ndez-Orallo \inst{2}\orcidID{0000-0001-9746-7632} 
}
\tocauthor{Manuel Garcia-Piqueras (\Letter)\inst{1}\orcidID{0000-0001-8088-8393} \&
Jos{\'e} Hern{\'a}ndez-Orallo \inst{2}\orcidID{0000-0001-9746-7632}
}

\authorrunning{Garcia-Piqueras and Hern{\'a}ndez-Orallo}

%
\institute{Math. Dept., Universidad de Castilla-La Mancha, Albacete, Spain\\
\email{manuel.gpiqueras@uclm.es}
\and
VRAIN, Universitat Polit{\`e}cnica de Val{\`e}ncia, Val{\`e}ncia, Spain\\
\email{jorallo@upv.es}
}

\maketitle              
\begin{abstract}
Recent research in machine teaching has explored the instruction of any concept expressed in a universal language. In this compositional context, new experimental results have shown that there exist data teaching sets surprisingly shorter than the concept description itself. However, there exists a bound for those remarkable experimental findings through {\em teaching size} and concept complexity that we further explore here. As concepts are rarely taught in isolation we investigate the {\em best} configuration of concepts to teach a given set of concepts, where those that have been acquired first can be reused for the description of new ones. This new notion of conditional teaching size uncovers new insights, such as the {\em interposition} phenomenon: certain prior knowledge generates simpler compatible concepts that increase the teaching size of the concept that we want to teach. This does not happen for conditional Kolmogorov complexity. Furthermore, we provide an algorithm that constructs {\em optimal curricula} based on interposition avoidance.  
This paper presents a series of theoretical results, including their proofs, and some directions for future work. 
New research possibilities in curriculum teaching in compositional scenarios are now wide open to exploration.

\end{abstract}

\section{Introduction}
\toctitle{Introduction}
\label{introduction}

Let us consider a {\em teacher} who instructs a given set of concepts by examples to a {\em learner}. Ideally, the teacher would design a curriculum such that the whole teaching session is shortest. 
For  one concept, the field of {\em machine teaching} has analysed the efficiency of the teacher, the learner or both, for different representation languages and teaching settings \cite{zhu2015,cicalese2020teaching,icml2020kumar,icml2020rakhsha,icml2020such}. 

{For more than one concept, however, we need to consider different sequences of examples, or {\em curricula}, to make learning more effective. While there has been extensive experimental work in curriculum learning \cite{soviany2021curriculum}, the theoretical analysis 
is not abundant and limited to continuous models 
\cite{Pentina_2015_CVPR,gong2016curriculum,weinshall2018}. It is not well understood how curriculum learning can be optimised when concepts are {\em compositional}, with the underlying representation mechanisms being rich languages, even Turing-complete. Also, in a curriculum learning situation where a {\em teacher} chooses the examples sequentially, it is surprising that the connection with machine teaching has not been made explicit at a general conceptual level, with only a specific minimax approach for gradient-based representations \cite{zhou2018minimax,gong2016multi,gong2017exploring,gong2019multi}
.   
In other words,} to our knowledge, a 
theoretical framework has not yet been articulated for curriculum learning in machine teaching, or {\em curriculum teaching}, when dealing with universal languages, as a counterpart to incremental inductive inference based on simplicity  \cite{solomonoff1964formal1,
solomonoff1989}.

While the teaching dimension has been the traditional metric for determining how easy it is to teach a concept \cite{zhu2015}, the {\em teaching size} \cite{Telle2019TheTS} is a new metric that is more reasonably related to how easy it is to teach an infinite compositional concept class. It is also more appropriate to understand `prompting' of language models as a kind of teaching, where users need to think of the shortest prompts that make a language model such as BERT, GPT-2 or GPT-3 achieve a task by few-shot learning \cite{devlin2018bert,radford2019language,brown2020language}.
%
%
However, as far as we know, 
the following issues are not clear yet: (1) {\em What is the relationship between the Kolmogorov complexity of a concept and how difficult it is to be taught under the teaching size paradigm?} and (2) {\em Is there a way to extend 
machine teaching, and teaching size in particular, to consider the notion of {\em optimal teaching curricula}?}

The first question prompts us to our first group of contributions: important theoretical results related to teaching size bounds. 
Theorem \ref{tsupperboundcomplexity} shows that 
 concepts with {\em high} complexity are {\em difficult to teach}, putting a limit to the surprising experimental finding recently reported in \cite{Telle2019TheTS}, where teaching a concept by examples was usually more economical (in total number of bits) than showing the shortest program for the concept. This connection between teaching size and complexity suggests that the second question may rely on a strong relation between incremental learning  using simplicity priors and curriculum teaching.  

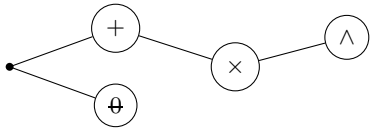
\begin{figure}[!h]
    \centering
    \begin{tikzpicture}[scale=1.5]
      \tikzstyle{every node}=[draw,shape=circle];
      \path (20:1cm) node (v1) {$+$};
      \path (0:2cm) node (v2) {$\times$};
      \path (5:3cm) node (v3) 
      {$\wedge$};
      \path (-20:1cm) node (v4) {\sout{0}};
        
      \fill (0,0) circle (1pt);
      \draw (0,0) -- (v1)
            (v1) -- (v2)
            (v2) -- (v3)
            (0,0) -- (v4);
            
    \end{tikzpicture}
    \vspace{-0.3cm}
    \caption{Curriculum teaching for a set of concepts.}
    \label{fig:curriculum}
\end{figure}

For instance, consider the concepts $c_{+}$ for addition, $c_{\times}$ for multiplication, $c_{\wedge}$ for  exponentiation and $c_{\stkout{0}}$ for the removal of zeros (Fig. \ref{fig:curriculum}). If  the concept of $c_+$ is useful to allow for a shorter description of $c_{\times}$, is it also reasonable to expect that $c_+$ would also be useful to {\em teach} $c_{\times}$ from examples? Or even $c_{\wedge}$? In general, is the conditional algorithmic complexity $K(c_2|c_1)$ related to the minimal size of the examples needed to teach $c_2$ after having acquired $c_1$?  

\vspace{-0.2cm}
\vspace{-0.3cm}

Our perspective studies the sequence of learning a set of concepts, instead of learning a sequence of instances under the same concept. 
In the general case, 
we define a teaching curriculum as a set of partial alternative sequences, such as the top and bottom branches in Fig. \ref{fig:curriculum}. The order between branches is irrelevant, but the order of concepts inside each branch is crucial. This tree structure is proposed as future work in 
\cite{Pentina_2015_CVPR}. 
Given a set of concepts, is there a curriculum that minimises the overall teaching size?  

Our second group of contributions turns around this new concept of teaching curriculum. We provide a definition of {\em conditional teaching size}, given some other concepts already taught, $TS(c|c_1, \dots, c_n)$. 
%
We show that, in general, 
$K( c_1|c_2) < K( c_2|c_1)$, for conditional Kolmogorov complexities, does not imply $\label{eq:2true}
TS(c_1|c_2) < TS(c_2|c_1)$, and vice versa.  
Furthermore, given a concept $c$, it is not true that $TS(c|B) \leq TS(c)$, $\forall B$. We find a 
new {\em interposition} phenomenon: acquired concepts may increase 
the teaching size of new concepts. 
We give conditions to avoid or provoke interposition. 
Theorems \ref{th_@} and \ref{th_@@} 
are key results in this direction, providing an 
explicit range 
where interposition might happen. 
Finally, we present an effective procedure, $\mathbb{I}$-{\em search}, to design {\em optimal} curricula, minimising overall teaching size, for a given set of concepts.

We will see these results in the following sections, 
and interpret their relevance in the context of other machine teaching paradigms. 
This opens up new avenues for understanding curriculum learning, and new strategies for defining incremental teaching protocols for several applications, especially in compositional scenarios and language model `prompting'.

\section{Notation and background}\label{section2}

%
%
Let us consider a machine $M$ and a universal (i.e., Turing complete) language $L$. 
We assume that $L$ is formed by a finite set of instructions in an alphabet $\Upsilon$,  
%
%
each of them been coded with the same number of bits. Hence, each program $p$ in language $L$ can simply be represented as a string in $\Sigma= \{0, 1\}^*$,  
whose length is denoted by  $\ninst(p)$ (in number of instructions) and denoted by $\ell(p)$ (in bits). 
%
There is a total order, $\prec$, over programs in language $L$ defined by two criteria: (i) length and (ii) lexicographic order over $\Upsilon$ only  applied  when two programs have equal size. 
%
Programs map binary strings in $\Sigma$ to ${\Sigma}\cup{\perp}$, denoted by $p(\mathtt{i})=\mathtt{o}$, with  
$p(\mathtt{i})=\perp$ representing that $p$ does not halt for $\mathtt{i}$. 
Two programs are equivalent if they compute the same function. 

We say that $c$ is an $L$-concept if it is a total or partial function 
$c:\Sigma \rightarrow {\Sigma}\cup{\perp}$   
computed by at least a program in language $L$. The class of concepts defined by all programs in $L$ is denoted by $C_L$; $[p]_L$ 
denotes the equivalence class of program $p$. Given $c \in C_L$, we denote $[c]_L$ as the equivalence class of programs in 
$L$ that compute the function defined by $c$. 
Examples are just pairs of strings, and their space is the infinite set $
X=\{ \langle \mathtt{i},\mathtt{o} \rangle : \langle \mathtt{i},\mathtt{o} \rangle \in {\Sigma} \times ({{\Sigma}\cup{\perp}}) \}$.
%
A {\em witness} can be any finite example subset of $X$, of the form $S=\{ \langle \mathtt{i}_1,\mathtt{o}_1 \rangle, \ldots, \langle \mathtt{i}_k,\mathtt{o}_k \rangle \}$. 
%
In order to calculate the {\em size} of these sets, we consider self-delimiting codes. 
Let $\delta$ 
be the number of bits needed to encode $S$, using certain prefix 
code. For instance, if we consider Elias coding \cite{eliascoding}, the string $01010010001001000101$ (size = $20$) expresses the example set 
$\{\langle 1, 010 \rangle, \langle 0, 1 \rangle\}$ unambiguously. The size of an example set is the size of its encoding (e.g., $\delta(\{\langle 1, 010 \rangle, \langle 0, 1 \rangle\}=20$ in Elias coding). For output strings, the natural number to be encoded is increased by 1, to accommodate for $\perp$. We also define a total order $\lessdot$ on $X$, 
i.e., $\forall S, S'$ such that $S \lessdot S'$ then $\delta(S) \leq \delta(S')$ with any preference (e.g., lexicographic) for equal size. 

A concept $c$ defines a unique subset of the example space $X$ and we call any element in that  subset a {\em positive} example. A concept $c$ satisfies example set $S$, denoted by $c \vDash S$, if $S$ is a subset of the positive examples of $c$.  
For instance, a witness set for the concept $c_{\stkout{0}}$ ({\em remove zeros}) is $\{\langle 10011, 111\rangle, \langle 001, 1\rangle\}$. Example sets cannot have different outputs for equal inputs: $\{\langle 1, 00\rangle, \langle 1, 01\rangle\}$ is not valid.

A program $p$ is compatible with $S=\{ \langle \mathtt{i}_j,\mathtt{o}_j \rangle\}^k_{j=1} \subset X$, denoted by $p \vDash S$, if $p_S(\mathtt{i}_j)=\mathtt{o}_j$ for every $j \in \{1, \ldots, k \}$. 
For a finite example set $S$, there is always a program, denoted by $\ddot{p_S}$, that implements a conditional hard-coded structure of if-then-elses (trie) specifically designed for $S$. 
If we know the number of bits of input $\mathtt{i}$ and the set of examples in $S$, the number of comparisons using a trie-data structure is linearly time-bounded. Namely, for any $\ddot{p_S}$, there exists a constant, $\rho$, such that $
    \rho \cdot \min\{\ell(\mathtt{i}), \ell(\mathtt{i}_{max})\} + \ell(\mathtt{o}_{max})$
is an upper bound of time steps for each input $\mathtt{i}$, 
where $\ell(\mathtt{i}_{max})$, $\ell(\mathtt{o}_{max})$ are the lengths of the longest input string and output string in $S$, respectively. 
%
%
%
In general, for any program that employs a trie-data structure for $S$, there exists a time-bound linear function, denoted by $\lambda_L(\mathtt{i}, S)$, that represents an upper bound in time steps 
on every input $\mathtt{i}$. 

Complexity functions $\mathsf{f}:\mathbb{N}\rightarrow\mathbb{N}$ act as time bounds. We say that a program $p$ is $\mathsf{f}$-compatible with the example set $S=\{ \langle \mathtt{i}_j,\mathtt{o}_j \rangle\}^k_{j=1} \subset X$, denoted by $p \vDash_{\mathsf{f}} S$, if $p(\mathtt{i}_j)=\mathtt{o}_j$ within $\max\{\mathsf{f}(\ell(\mathtt{i}_j)), \lambda_L(\mathtt{i}_j, S)\}$ time steps (time-bound) for each $j \in \{1, \ldots, k\}$. In other words, within time bound, for each pair $\langle \mathtt{i}, \mathtt{o} \rangle \in S$ the program $p$ on input $\mathtt{i}$: (1) outputs $\mathtt{o}$  when $\mathtt{o} \neq \perp$ or (2) does not halt when $\mathtt{o} = \perp$. 
%
%
Note that: (i) For any complexity function $\mathsf{f}$ and any example set $S$, there is always\footnote{Note that this $\ddot{p}_S$ is ensured by the $\max$ with time costs.}, 
a program $\mathsf{f}$-compatible with $S$, (ii) there may be  programs $p$ such that $p \nvDash_\mathsf{f} S \wedge p \vDash S$, if $\mathsf{f}$ and $S$ do not guarantee enough time bound 
and (iii) larger complexity functions distinguish more programs. 


\section{Absolute teaching size and complexity}\label{section3}

Now we can study how a non-incremental teacher-learner setting works and the relationship between teaching size and Kolmogorov complexity. 

Following the teaching settings based on the K-dimension \cite{balbach2007models,balbach2008measuring}, seen as preference-based teaching  using simplicity priors \cite{gao2017preference,hernandezfinite}, we assume that the learner is determined to find the shortest program (according to the prior $\prec$). Namely, the learner $\Phi$ returns the first program, in order $\prec$, for an example set $S$ and a complexity function $\mathsf{f}$ as follows: 
\begin{equation*}
    \Phi_{\ell}^\mathsf{f}(S) = {\arg \min \limits_{p}}^{\prec} \left\{  \ell (p) : p \vDash_\mathsf{f} S  \right\}
\end{equation*}
Note that the $\mathsf{f}$-bounded Kolmogorov complexity of an example set $S$, $K^\mathsf{f}(S)$, is the length of the program returned by the learner $K^\mathsf{f}(S) = \ell(\Phi_{\ell}^\mathsf{f}(S))$. We say that $S$ is a {\em witness set} of concept $c$ for learner $\Phi$ if $S$ is a finite example set such that $p = \Phi_{\ell}^\mathsf{f}(S)$ and $p \in [c]_L$. 

The teacher selects the {\em simplest} witness set that allows the learner to identify the concept, according to set size ($\delta$) and associated total order $\lessdot$, 
as follows:
\begin{equation*}
    \Omega_{\ell}^\mathsf{f}(c) = {\arg \min \limits_{S}}^{\lessdot} \left\{  \delta(S) : \Phi_{\ell}^\mathsf{f}(S) \in [c]_L) \right\}
\end{equation*}
The $K^\mathsf{f}$-teaching size of a concept $c$ is $TS_{\ell}^\mathsf{f}(c) = \delta(\Omega_{\ell}^\mathsf{f}(c))$.


Every program the teacher picks defines a concept $c$.  
The teacher-learner protocol is computable for any complexity function $\mathsf{f}$ and able to create pairs $(p_c, w_c)$, where $p_c$ defines a concept $c$ and $w_c$ is a witness set of $c$. We can think of these pairs as if they were inserted sequentially in the so-called {\em $\mathsf{f}$-Teaching Book} ordered by $w_c$, with no repeated programs or witness sets. 
%
For example, if we consider the concept $a \in C_L$ for swapping ones and zeros in a binary string, there will be a pair
$(p_a, w_a)$ in the $\mathsf{f}$-Teaching Book, e.g., containing a witness set like  $w_a = \{\langle 10, 01 \rangle, \langle 110, 001 \rangle \}$ that the teacher would provide with which the learner would output $p_a$, a program that swaps $1$ and $0$. 
%
Theorem 1 in \cite{Telle2019TheTS} shows that {\em for any concept $c \in C_L$, there exists a complexity function $\mathsf{f}$ such that there is a pair $(p_c, w_c)$ in the $\mathsf{f}$-Teaching Book}. 
The teaching size makes more sense than the traditional teaching dimension (the smallest cardinality of a witness set for the concept) because some concepts could be taught by very few examples, but some of them could be extremely large. Also, the use of size instead of cardinality allows us to connect teaching size and Kolmogorov complexity, as we do next.


Our first result shows an equipoise between teaching size and data compression, an extra support for machine teaching; the compressing performance of the learner and the minimisation of the teaching size go in parallel. 

\begin{proposition}
\label{TSComplexityBounds}
{\em Let $\mathsf{f}$ be a complexity function and $\Phi_{\ell}^\mathsf{f}$  the learner.  
There exist two 
constants $k_1, k_2 \in \mathbb{N}$, such that for any given pair $(w, p) \in \mathsf{f}$-Teaching Book we have that:}\footnote{We use the standard definition of $K$ using a monotone universal machine $U$ \cite{LiVitanyi} (we will drop $U$ when the result is valid for any $U$), applied to binary strings (where programs and example sets are encoded as explained in the previous section). With $K^f$ we 
refer to a non-universal version where the descriptional machine is the learner.}
%
\begin{equation}\label{eq:upperboundcomplexity}
    K(p) \leq \delta(w) + k_1 \textit{ and } K(w) \leq \ell(p) + k_2
\end{equation}
\begin{proof}
To begin with, we address the first part of the statement \ref{eq:upperboundcomplexity}. 
Recall that $\Phi_{\ell}^\mathsf{f}$ is a one-to-one computable function:
\begin{align*}
  \Phi_{\ell}^\mathsf{f} \colon &\Sigma \to \Sigma\\
  & w \mapsto p.
\end{align*}

It is important to state that the domain of this function is the set of sets of examples $w$ included in the $\mathsf{f}$-Teaching Book. In addition, this function is computable since every pair $(w, p) \in \mathsf{f}$-Teaching Book is obtained through the $\mathsf{f}-$bounded \textit{learner} \cite{Telle2019TheTS}. 

Let us recall the invariance theorem, for any two machines (i.e., description modes) $U,V$, with $U$ universal, there is a constant $k \in \mathbb{N}$ such that for every $s$ we have:
\begin{equation}\label{eq:optimaldescriptionmode}
    K_U(s) \leq K_{V}(s) + k
\end{equation}


Interestingly, we can use $\Phi_{\ell}^\mathsf{f}$ as a description mode (a machine $V$), since it is a one-to-one function that goes from binary strings to binary strings. We can then define the Kolmogorov complexity of $p$, with respect to the description mode $\Phi_{\ell}^\mathsf{f}$ (which will be referred from now on simply as $\Phi$):
\begin{equation}\label{eq:complexityphi}
    K_{\Phi} (p) = min \{ \delta(w): \Phi(w)=p \}
\end{equation}

It follows, from inequality \ref{eq:optimaldescriptionmode} and definition \ref{eq:complexityphi}, 
that for every $U$:
\begin{equation}\label{eq:optimaldescriptionmodephi}
    K_U(p) \leq K_\Phi(p) + k_1
\end{equation}

\noindent Since $K_{\Phi}(p) = \delta(w)$ as $(w, p)$ is a pair of the $\mathsf{f}$-Teaching Book. Then, the inequality \ref{eq:optimaldescriptionmodephi} can be expressed as:
\begin{equation*}
    K(p) \leq \delta(w) + k_1
\end{equation*}

\noindent 
Note that $U$ has been dropped as this is valid for any universal machine $U$.

Now we address the second part of statement \ref{eq:upperboundcomplexity}. We recall that every pair $(p, w)\in \mathsf{f}$-Teaching Book is unique, i.e., there is just one and only one $p \in \Sigma$ such that $\Phi(w)=p$. In other words, the function $\Phi \colon \Sigma \to \Sigma$ 
%
%
is one-to-one. Recall also that this function is defined over the set of examples $w$ of the $\mathsf{f}$-Teaching Book. So that, there is an inverse (partial)\footnote{A partial function, as the original $\Phi$ was.} function $\Phi^{-1} \colon \Sigma \to \Sigma$.
%

Clearly, $\Phi^{-1}$ is also a computable function 
just by looking into the book. 
Now, we are able to consider $\Phi^{-1}$ as a description mode (a machine $V$) and we can obtain the Kolmogorov complexity of $w$ with respect to $\Phi^{-1}$ as:
\begin{equation}\label{eq:complexityphi_inverse}
    K_{\Phi^{-1}}(w) = min \{\ell(p): \Phi^{-1}(p)=w\}
\end{equation}
\noindent Again, since $(p, w)$ is a pair of the $f$-teaching book then $K_{\Phi^{-1}}(w) = \ell(p)$. Similarly to the reasoning already used to obtain the first part of statement \ref{eq:upperboundcomplexity}, through inequality  \ref{eq:optimaldescriptionmode} and definition \ref{eq:complexityphi_inverse}, there exists a constant $k_2$, which does not depend on $p$, such that
\begin{equation*}
    K(w) \leq \ell(p) + k_2
\end{equation*}

\end{proof}
\end{proposition}
%
%
%
Proposition \ref{TSComplexityBounds} is a key result ensuring that the size difference between programs and witness sets is bounded: a short witness set would not correspond with an arbitrarily complex concept and vice versa. This puts a limit to the surprising empirical observation in \cite{Telle2019TheTS}, where the size of the witness sets in bits was usually smaller than the size of the shortest program for that set, i.e., in terms of information it was usually cheaper to teach by example than sending the shortest description for a concept.

There is another close relationship between the Kolmogorov complexity of a concept and its teaching size. 
First we need to define the complexity of a concept through the {\em first program} of a concept in language $L$.
%
\begin{equation*}
    p_c^*={\arg \min \limits_{p}}^{\prec} \left\{  \ell(p) : p \in [c]_L \right\}
\end{equation*}
%
%
%
For every concept $c \in C_L$, we will simply refer to the 
Kolmogorov
 complexity of a concept $c$ with respect to the universal language $L$ 
 as $K_{L}(c)=\ell(p_c^*)$.
Now, 
%
\begin{theorem}
\label{tsupperboundcomplexity}{\em Let $L$ be a universal language, $M$ be a universal machine and $k_M$ be a constant that denotes the length of a  program for $\Phi$ in $M$.\footnote{For any universal Turing machine $M$, a finite program can be built coding an interpreter for $\Phi$ in $M$ and taking $w_c$ as input. The length of this `glued' program does not depend on the concept $c$ but on the machine $M$ to glue things together and how many bits of the program instructions are required to code $\Phi$, i.e., $K_M(\Phi)$.} For any concept $c \in C_L$, there exists a complexity function $\mathsf{f}$, such that $K_{L}(c) \leq TS_{\ell}^\mathsf{f}(c)+k_M$.}
\begin{proof}
Theorem 1 \cite{Telle2019TheTS} guarantees the existence of a complexity function $\mathsf{f}$ and a witness set, $w_c$, such that the learner, $\Phi_{\ell}^\mathsf{f}$, outputs $p_c$, on input $w_c$, satisfying: \begin{equation*}
    p_c \in [c]_L
\end{equation*}

Then, $p_c$ computes the same partial function that is defined by $c$. In other words, $p_c$ is a description of $c$ procured through the witness set $w_c$, the learner, $\Phi_{\ell}^\mathsf{f}$, and 
some {\em glue} program of size $\epsilon$ that
executes $\Phi_{\ell}^\mathsf{f}$ on input $w_c$. 
As a result, $K_L(c)$ cannot be greater than the addition of $\delta(w_c)$, $K_M(\Phi_{\ell}^\mathsf{f})$ and $\epsilon$, i.e., 
\begin{equation*}
    K_L(c) \leq TS_{\ell}^\mathsf{f}(c)+K_M(\Phi_{\ell}^\mathsf{f})+\epsilon
\end{equation*}
\end{proof}
\end{theorem}
\noindent This gives an upper bound (the teaching size) for the Kolmogorov complexity of a concept. On the other hand, this theorem implies that concepts with {\em high} complexity are {\em difficult to teach} in this setting. The surprising observation found in \cite{Telle2019TheTS} of some concepts having shorter TS than K has a limit.


\section{Conditional teaching size}\label{section4}
In this section we introduce the notion of conditional teaching size and the {\em curriculum teaching problem}. We now assume that the learner can reuse any already learnt concept to {\em compose} other concepts. The curriculum teaching problem is  
to determine the optimal {\em sequential} way of teaching a set of concepts $Q=\{ c_1, c_2, \, \dots, c_n \}$, in terms of minimum total teaching size. 
%
Let $TS(c_i|c_j, c_k \, \dots)$ be the conditional teaching size of concept $c_i$, given the set of concepts $\{ c_j, c_k \, \dots \}$ previously distinguished by the learner. The challenge is to minimise
$TS(c_1) + TS(c_2|c_1) + TS(c_3|c_1, c_2) + \dots$.

In this new setting we need a definition of $TS(c_i|c_j)$ that considers that (1) a concept $c$ has infinitely many programs that generate it, so which one the learner has identified may be important, and (2) the learner must have some {\em memory}, where that program is stored. 
Interestingly, if we assume that memory is implemented by storing the identified programs in a library, where the learner can only make calls to ---but not reuse its parts---, then it is irrelevant which program has been used to capture concept $c$, since the learner only reuses the functional {\em behaviour} of the program\footnote{Note that the learner may use a complexity function $\mathsf{f}$. If that is the case, it can occur that a particular program $p_1$ identifies $c_1$ and $c_1$ is very useful for $c_2$, but $p_1$ is too slow to be used in any reasonably efficient program for $c_2$, so becoming useless incrementally. 
The computational time of the learner has also been considered in other machine teaching frameworks \cite{10.5555/3305890.3305903,zhu2018overview}.}.

\subsection{Conditional teaching size and minimal curriculum}
We define a library $B=\{ p_1, \dots, p_k \}$, as a set of programs in the universal language used by the learner. Let $|B| = k$ the number of {\em primitives}.  
%
%
We assume that $\Upsilon$ always includes an instruction $@$ for making static\footnote{There is no loss of generality here, since every program that uses dynamic calls can be rewritten only using static calls \cite{1235426}.} library calls. 
We use $@\mathsf{i}$ to denote the instruction that calls the primitive that is indexed as $\mathsf{i}$ in the library. If $|B|=1$, then $@$ needs no index. 
%
%
%
Accordingly, the length of a call to the library is $\ell(@\mathsf{i})=\ell(@)+\log_2(|B|)$ $=\log_2(|\Upsilon|)+ \log_2(|B|)$ bits.

Let $p$, $p'$ be programs in the universal language $L$ and $B$ a library. We say that a program $p$ {\em contains a  call to} $p'$ when
$@\mathsf{i}$ is a substring of $p$
and 
$\mathsf{i}$ is the index of $p' \in B$. 
%
$L_{B}$ denotes a language $L$ that implements static calls to a library $B$. 
Even with static calls, the flow of the program may never reach $@$ for an input. 
Interestingly, we can avoid this undecidable question when dealing with programs in the teaching book by considering $@$ as the last instruction regarding lexicographical order.

\begin{lemma}
{\em Let $\mathsf{f}$ be a complexity function and $B$ a library. For any $(w, p)$ $\in$ $\mathsf{f}$-Teaching Book, if $p$ has a call to $B$ then $p$ effectively reaches $@$ and executes a primitive on at least one input of $w$.}\sidenoteJ{We may need to assume that there is a NOP instruction, so that there is a program that precedes @xx.. and replaces it by NOP NOP NOP.. Pero esto se puede decir si hace falta en la demo del lema.}\sidenoteM{No sabía lo que era una NOP instruction. Ahora que lo he mirado no entiendo muy bien lo que quieres decir.}\sidenoteJ{Es una instrucción que no hace nada. En ensamblador antiguo solía estar porque venía bien a veces. En nuestro caso nos podría valer para decir que @ siempre se puede sustituir por NOP, pero mejor olvidarlo si no hace falta.}
\begin{proof}
If the learner identifies a program $p$ that incorporates $@$ without executing it for any example in a given witness set $w$, there must be a previous program, smaller in size, that does not call the library. The only issue of this rationale is that there are some languages that {\em skip} some instructions, which are never executed. But, in this case, $p$ will not include such instructions, because the learner would have identified a shorter program without them.
\end{proof}
\end{lemma}

Let us use $\dot{p}$ to denote  program $@\mathsf{i}$, where $\mathsf{i}$ is the index of $p$ in the library.

\begin{lemma}\label{irreducibility}
{\em Let $B$ be a library. The language $L_B$ satisfies: $\dot{p} \prec p', \,\, \forall p'\text{ such that } p' \notin [p]_L \wedge p' \text{ has a call to } p$.}
\begin{proof}
Consider a program $p'$ such that: $p' \notin [p]_L$ and $p'$ {\em calls} $p$. If $p'$ has a {\em call} to $p$ then $@\mathsf{i}$ is a substring of $p'$, where $\mathsf{i}$ points to the primitive $p$. Either $p'$ is $\dot{p}$ or is larger; in the latter case it is posterior in the order $\prec$. 
\end{proof}
\end{lemma}

Now, we are able to redefine the learner, $ \Phi_{\ell}^\mathsf{f}$, and the time-bounded Kolmogorov complexity for a given library. 

\begin{definition}
{\em Let $\mathsf{f}$ be a complexity function, $B$ a library and $S$ an example set. The learner $\Phi$ calculates the {\em first program} for $S$ in language $L_B$:}
\begin{equation*}
    \Phi_{\ell}^\mathsf{f}(S|B) = {\arg \min \limits_{p \in L_B}}^{\prec} \left\{  \ell (p) : p \vDash_\mathsf{f} S  \right\}
\end{equation*}
{\em The $\mathsf{f}$-bounded Kolmogorov complexity of $S$, denoted by $K^\mathsf{f}(S|B)$, is the length of the program returned by the learner:}
$K^\mathsf{f}(S|B) = \ell \left( \Phi_{\ell}^\mathsf{f}(S|B) \right)$. 
{\em The extension of the teacher, denoted by $ \Omega_{\ell}^\mathsf{f}(c|B)$, also selects the shortest witness set that makes the learner distinguish the concept:}
\begin{equation*}
    \Omega_{\ell}^\mathsf{f}(c|B) = {\arg \min \limits_{S}}^\lessdot \left\{  \delta(S) : \Phi_{\ell}^\mathsf{f}(S|B) \in [c]_{L_{B}}) \right\}
\end{equation*}
{\em And the definition of the $K^\mathsf{f}$-teaching size of a concept $c$ is $TS^\mathsf{f}_\ell (c|B) = \delta(\Omega_{\ell}^\mathsf{f}(c|B))$.}
\end{definition}



We can also extend Theorem 1 in \cite{Telle2019TheTS}.

\begin{corollary}\label{Theorem1extended}
{\em Let $L$ be a universal language and $B$ a library. For any concept $c$ in $C_{L_B}$, there is a complexity function $\mathsf{f}$ so that the $\mathsf{f}$-Teaching Book will contain some $(p_c, w_c)$ with $p_c \in [c]_{L_B}$ and $TS_{\ell}^\mathsf{f}(c|B) = \delta(w_c)$.}
\begin{proof}
The result is direct since $L_B$ is a universal language and we can apply Theorem 1 in \cite{Telle2019TheTS} directly.
\end{proof}
\end{corollary}

Sometimes we will refer to the original $L$ and sometimes to the augmented $L_B$ depending on whether we see it conditional to $B$ or not. 

We are now in position to give a formal definition of the conditional teaching size given a set of concepts.

\begin{definition}
{\em Let $a \in C_L$, $\{c_i\}_{i=1}^n \subset C_L$ and let $p_i = \Phi(\Omega_{\ell}^\mathsf{f}(c_i))$, for each $i=1, \dots, n$. Let $B=\{p_i\}_{i=1}^n$. We define the conditional teaching size of concept $a$ given the concepts $\{c_i\}_{i=1}^n$, denoted by $TS_{\ell}^\mathsf{f}(a|c_1, \dots, c_n)$, as}
\begin{equation*}
   TS_{\ell}^\mathsf{f}(a|c_1, \dots, c_n)=TS_{\ell}^\mathsf{f}(a|B) 
\end{equation*}
\end{definition}
The programs that identify the concepts are in the same 
$\mathsf{f}$-Teaching Book.

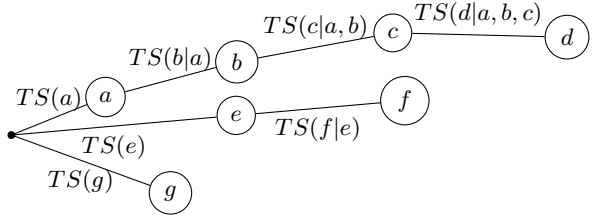
\begin{figure}[h!]
    \centering
    \begin{tikzpicture}[scale=1.5]
      \path (22:0.9cm) node[draw,shape=circle] (v0) {$a$};
      \path (18:2.1cm) node[draw,shape=circle] (v1) {$b$};
      \path (15:3.5cm) node[draw,shape=circle] (v2) {$c$};
      \path (10:5cm) node[draw,shape=circle] (v3) {$d$};
      \path (5:2cm) node[draw,shape=circle] (v4) {$e$};
      \path (5:3.5cm) node[draw,shape=circle] (v5) {$f$};
      \path (-20:1.5cm) node[draw,shape=circle] (v6) {$g$};
        
      \fill (0,0) circle (1pt);
      \draw (0,0) -- node[above]{$TS(a)$}(v0)
            (v0) -- node[above]{$TS(b|a)$}(v1)
            (v1) -- node[above]{$TS(c|a, b)$}(v2)
            (v2) -- node[above]{$TS(d|a, b, c)$}(v3)
            (0,0) -- node[below]{$TS(e)$}(v4)
            (v4) -- node[below]{$TS(f|e)$}(v5)
            (0, 0) -- node[below]{$TS(g)$}(v6);
    \end{tikzpicture}
    \vspace{-0.2cm}
    \caption{Curriculum $\{a \rightarrow b \rightarrow c \rightarrow d,e \rightarrow f, g\}$ for a set of concepts $\{a, b, c, d, e, f, g\}$.}
    \label{fig:distribution}
\end{figure}

We now give a definition of {\em curriculum}. 
Given a set of concepts, a curriculum is a set of disjoint sequences covering all the concepts. 
Our notion of curriculum is more general than just a simple sequence.  
If some branches are unrelated, a curriculum should not specify which branch comes first, and are considered independent `lessons'. We will see how this flexibility is handled by the algorithm that finds the optimal curriculum in section \ref{section5}. 
For instance,  Fig. \ref{fig:distribution} shows how a set of concepts $\{a, b, c, d, e, f, g\}$ is partitioned into three branches: $\{a \rightarrow b \rightarrow c \rightarrow d , e \rightarrow f , g\}$, where $a \rightarrow b$ means that $b$ must come after $a$ in the curriculum. For each {\em branch}, there is no background knowledge or library at the beginning. The library grows as the teacher-learner protocol progresses in each branch.

\vspace{-0.5cm}
\vspace{-0.8cm}


\begin{definition}
{\em Let 
$Q=\{c_i\}_{i=1}^n$ a set of $n$ labelled concepts. A curriculum $\pi = \{ \sigma_1, \sigma_2, \cdots, \sigma_m \}$ is a full partition of $Q$ where each of the $m$ subsets $\sigma_j \subset Q$ has a total order, becoming a sequence. 
Also, we denote $\overline{Q}$ as the set of all the curricula in $Q$.}
\end{definition}

The order in which the subsets are chosen does not matter, but the order each subset is traversed does. For example, the curriculum $\pi = \{a \rightarrow b \rightarrow c \rightarrow d , e \rightarrow f , g\}$ can have many paths, such as $abcdedfg$ or $gabcdef$. But note that $\pi$ is different from $\pi'=\{b \rightarrow a \rightarrow c \rightarrow d , f \rightarrow e , g \}$. 
For any $Q$ with $n$ concepts, the number of different curricula is \begin{equation}\label{number_curricula}
    |\overline{Q}| = n! \cdot \Biggl( \sum_{k=0}^{n-1} \binom{n-1}{k} \cdot {{1}\over{(k+1)!}} \Biggr)
\end{equation}

Let us explain how to get such number of distinct curricula (Eq. \ref{number_curricula}), since the rationale is useful to generate them. {\em For any set $Q=\{c_1, \dots, c_n\}$ of $n$ 
concepts, the total number of different curricula is: }

There are $n!$ permutations of $n$ labelled elements. For each permutation, there are $n-1$ possibilities of starting a {\em branch}. So that, we can choose $k$ positions out of $n-1$. It implies that there will be $k+1$ subsets which can change its order, i.e., $(k+1)!$ different permutations of the subsets express the same case.
Therefore, there are $n! \cdot \binom{n-1}{k} \cdot {{1}\over{(k+1)!}}$ cases. Since $k \in \{0, 1, \ldots, n-1\}$, the Eq. \ref{number_curricula} gives the total number of distinct curricula.

In what follows we will consider that the concepts we work with are all in the original 
$\mathsf{f}$-Teaching Book, so they can be taught independently. This is not an important constraint, given Theorem 1 in \cite{Telle2019TheTS} and Corollary \ref{Theorem1extended} here. With this we ensure 
the same $\mathsf{f}$ for all of them. Now we can properly define the teaching size of a curriculum: 

\begin{definition}
Let $\mathsf{f}$ be a complexity function and let 
$Q$ be a set of concepts that appear in the original $\mathsf{f}$-Teaching Book. Let $\pi= \{ \sigma_1, \sigma_2, \cdots, \sigma_m \}$ a curriculum in $Q$. 
%
We define the teaching size of each sequence $\sigma = \{ c_1, c_2, ..., c_k \}$ as $TS_{\ell}^\mathsf{f}(\sigma) = TS_{\ell}^\mathsf{f}(c_1) + \sum_{j=2}^{k} TS_{\ell}^\mathsf{f}(c_j|c_1, \ldots, c_{j-1})$. 
The overall teaching size of $\pi$ 
is just  $TS_{\ell}^\mathsf{f}(\pi) = \sum_{i=1}^{m} TS_{\ell}^\mathsf{f}(\sigma_i)$.
\end{definition}

We say that a curriculum in $Q$ is minimal, denoted by $\pi^*$, if no other has less overall teaching size. But {\em how can we identify minimal curricula?} That is what we analyse next. 

\subsection{Interposition and non-monotonicity}

We now show a teaching phenomenon called {\em interposition}: new acquired concepts may lead to an increase in teaching size. The phenomenon might not even preserve the relationship established between two concepts, in terms of conditional Kolmogorov complexity, when considering conditional teaching size.

\begin{definition}
{\em We say that $B$ is an {\em interposed} library for concept $c$ if $TS(c|B)>TS(c)$; if $B=\{p'\}$ we say that $p'$ is an interposed program for $c$.}
\end{definition} 

\begin{proposition}
{\em For any $(w_c, p_c) \in \mathsf{f}$-Teaching Book, such that $@ \prec p_c$, there is an interposed library for concept $c$.}
\begin{proof}
Let $S=\{ \langle \mathtt{i}, \mathtt{o} \rangle \in w: \delta(w) \leq \delta(w_c) \wedge$ $c \vDash w\}$, i.e., the union of all the witness sets of length less or equal than $\delta(w_c)$ that are compatible with concept $c$.%

Let $\langle \mathtt{i}, \mathtt{o} \rangle$ be an input-output pair, which is not compatible with $c$ and $\mathtt{i}$ is not an input in $S$.

We now define $S'=S \cup \{\langle \mathtt{i}, \mathtt{o} \rangle\}$ as a trie program implementing a lookup table for $S'$. Let $B=\{\ddot{p_{S'}}\}$ be a library. For every witness set $w$, such that $\delta(w) \leq TS(c)$, it occurs $\Phi(w|B)=@ \notin [c]_{L_B}$, because $\ddot{p_{S'}}$ covers $\{\langle \mathtt{i}, \mathtt{o} \rangle\}$, but $c$ does not cover it. Therefore, any witness set $w'$, such that $\Phi(w'|B) \in [c]_{L_B}$, satisfies $TS(c)<\delta(w')$.
\end{proof}
\end{proposition}


The above proposition means that virtually every concept (the only condition is that is represented in the teaching book by a program of more than one instruction) may be interposed by a primitive that makes the witness set lead to another concept. This is an important result, as it is not only the case that some concepts may be useless for the concepts yet to come in the curriculum, but that they may even be harmful. This will have important implications when we look for minimal curricula in the following section.

This contrasts with conditional Kolmogorov complexity, where for every $a$ and $b$ we have that $K(a|b) \leq K(a)$. Given this, we can study the monotonicity between concept complexity and teaching size. Namely, {\em is there any relationship between $K(a|b) \leq K(b|a)$ and $TS(a|b) \leq TS(b|a)$?} 
We now show that, for any universal language, the inequalities aforementioned have, in general, different directions. First, we give the following definition.
\begin{definition}
{\em Let $c \in C_L$ and let $B$ be a library. We define the Kolmogorov conditional complexity of a concept $c$ given a library $B$ as $K_{L_B}(c) = \ell(p_c^*)$ where $p^*_c$ is calculated using $L_B$. We 
use the notation $K(c|B)=K_{L_B}(c)$.}
\end{definition}
We now extend the conditional Kolmogorov complexity to a set of concepts through programs that identify the concepts given in the same $\mathsf{f}$-Teaching Book and provide the result.
\begin{definition}
{\em Let $a \in C_L$, the set $\{c_i\}_{i=1}^n \subset C_L$ and $p_i = \Phi(\Omega_{\ell}^\mathsf{f}(c_i))$, for each $i=1, \dots, n$. Let $B=\{p_i\}_{i=1}^n$. We define the Kolmogorov complexity of concept $a$ given the concepts $\{c_i\}_{i=1}^n$, denoted by $K(a|c_1, \dots, c_n)$, as}
\begin{equation*}
   K(a|c_1, \dots, c_n) = K(a|B) 
\end{equation*}
\end{definition}
In words, the conditional complexity of a concept given a set of concepts  is equal to the conditional complexity of the concept given the canonical programs for those concepts as  extracted from the original teaching book. 

We now show the non-monotonicity between $K$ and $TS$:
\begin{theorem}\label{KTSInequalityChangeDirection}
{\em There exist two concepts $a$, $b\in C_L$ and a complexity function, $\mathsf{f}$, such that $K(a|b) < K(b|a)$ and $TS_{\ell}^\mathsf{f}(a|b) > TS_{\ell}^\mathsf{f}(b|a)$.}
\begin{proof}
In this proof we use a very unusual way of inputs which involve certain fixed `structure', which we use to define the behaviour of the concepts $a$ and $b$ we want to find. The structure can be used, for instance, to link it with the teaching size of a concept and it also helps to define a partial concept, as we will see below. 

Let us consider inputs in the form $\mathtt{i} = \mathtt{xyi'}$, which have three parts: 
\begin{itemize}
    \item $\mathtt{x}$ is a binary representation of a decimal number $n$ using $k$-digit codification, with $k \geq 2$.
    \item $\mathtt{y}$ is a binary string that concatenates, $k$ times, the binary string $01$.
    \item $\mathtt{i'}$ is a binary string which employs $k$-digit codification.
\end{itemize}

The purpose of such convoluted disposition is to have some redundancy in the coding, which we will exploit for the purposes of this proof. This kind of structure will be used to define very particular concepts that will display the inequalities in the theorem. 

For example, 
$\mathtt{i} = \overline{110011}0101\overline{\overline{1100110011}}$ expresses $10101$:
\begin{itemize}
\item The substring where there is no repetition of bits is $\mathtt{y}=0101$, then $k=2$, because $01$ appears twice.
\item The first part is $\mathtt{x}=\overline{110011}_{(2}$. Undoing the $k$-digit codification, $\mathtt{x}$ represents the binary string $101$, which gives the decimal number $n=5_{(10}$.
\item Since $n=5$, we take the last $k \cdot n =10$ binary digits, $\overline{\overline{1100110011}}$, as $\mathtt{i'}=\overline{\overline{1100110011}}$.
\item Now, undoing the $k$-digit codification, we are {\em confident} that $\mathtt{i}$ expresses the binary string $10101$ as input. 
\end{itemize}

And the other way round, for instance, if we want to express the input $1100$ through the structure $\mathtt{xyi'}$, then it is  uniquely expressed, in $k=2$ digit codification, as $\mathtt{i} = \overline{110000}0101\overline{\overline{11110000}}$.



We do not allow inputs starting with zeros. For instance, values of $\mathtt{x}$ such as $011$ or $001$ are not valid; we should write $11$ and $1$, respectively.

Note that we can rewrite every input, which do not start with zero, in the form $\mathtt{xyi'}$ and vice-versa. 


Let $b$ be a concept with very {\em high} complexity. It also needs every bit of the input, 
for instance, consider the class of programs that changes ones by zeros and vice-versa; every bit of the input is needed. If the input has another structure, different from $\mathtt{xyi'}$, it outputs $\perp$. 

Let $p_{\perp}$ be a program that outputs $\perp$ on every input. Concept $b$ shall satisfy $K(b)>K([p_{\perp}])$, since the complexity of $b$ must be very {\em high}. 

Theorem 1 in \cite{Telle2019TheTS}, guarantees the existence of a complexity function, $\mathsf{f}$, such that we can teach concept $b$. Let $(w_b, p_b) \in \mathsf{f}$-Teaching Book, with $TS_{\ell}^{\mathsf{f}}(b)=\delta(w_b)$.

We now define the program $p_a$ as $p_a(\mathtt{xyi'})=p_b(\mathtt{xyi'})$, when $\mathtt{x}$ represents a decimal number $n$, such that $n >\delta(w_b)$. Otherwise, $p_a(\mathtt{xyi'})=\bot$.


If the complexity of $b$ is high enough, then it is larger $K(b|a)$ than $K(a|b)$, because $a$ is a partial concept of $b$. So that, 
\begin{equation}\label{kinequality}
 K(a|b) < K(b|a)   
\end{equation}

We address now the teaching size of concepts $a$ and $b$. 

Suppose that $\mathsf{f}$ does not allow to teach concept $a$. Then, Theorem 1 in \cite{Telle2019TheTS} guarantees the existence of another complexity function to teach concept $a$. We set $\mathsf{f}$ as the greatest of both complexity functions.

We now take notice of $TS_{\ell}^{\mathsf{f}}(b|a)$. If the learner receives an input equal or smaller than $\delta(w_b)$, the library $B=\{p_a\}$ does not help to identify concept $b$. The only rationale against this is that $a$ may help to identify other programs. However, we can always increase the $k$-digit codification. If $k$ is high enough, then it is more difficult to produce an adequate input for $p_a$, so that, it will be shorter making $p_b^*$ from scratch, than using the primitive $p_a$. Therefore, 
\begin{equation}\label{tsba}
    TS^{\mathsf{f}}_{\ell}(b|a) \leq \delta(w_b)=n
\end{equation}

We now consider $TS^{\mathsf{f}}_{\ell}(a|b)$. We should need example sets larger than $n$ to identify concept $a$, but we could identify it considering witness sets having only $\perp$ as outputs. But this case is not possible. Since $K(a)>K(b)>K([p_{\perp}])$, it implies that $[p_{\perp}]$ precedes $a$ as an output of the learner when it gets, as input, any $w$ with $\delta(w)<n$. Therefore,

\begin{equation}\label{tsab}
    n < TS^{\mathsf{f}}_{\ell}(a|b)
\end{equation}

If we consider the inequalities \ref{tsba} and \ref{tsab}, then we get 
\begin{equation}\label{tsinequality}
    TS^{\mathsf{f}}_{\ell}(b|a) \leq n < TS^{\mathsf{f}}_{\ell}(a|b)
\end{equation}

The resolution follows from inequalities \ref{kinequality} and \ref{tsinequality}.
\end{proof}
\end{theorem}

When considering conditional teaching size for curriculum learning, we need general conditions to avoid interposition. For instance, an important reduction of program size in language $L_B$ usually minimises the risk of interposition.

\begin{corollary}\label{corol:strictdecreaseTS}
{\em Let $(w_c, p_c) \in \mathsf{f}$-Teaching Book, with $p_c \in [c]_L$. 
If there exists a library $B$ and a witness set $w$, verifying the following conditions (1) $\delta(w) < \delta(w_c)$ and (2) the first program $p_c' \in [c]_{L_B}$, using order $\prec$, such that $p_c' \vDash_{\mathsf{f}} w$, precedes any other program $p$ in language $L_B$, satisfying $p \vDash_\mathsf{f} w$, then $TS^{\mathsf{f}}_{\ell}(c|B)< TS^{\mathsf{f}}_{\ell}(c)$.}
\begin{proof}
If $\Phi(w'|B) \in [c]_{L_B}$, for certain $w'$ such that $w' \lessdot w$, then we have the conclusion. Otherwise, the learner gets $w$ as input; thus, (2) guarantees \begin{equation*}
    \Phi(w|B) \in [c]_{L_B}
\end{equation*}
\end{proof}
\end{corollary}


These conditions to avoid interposition are quite strong, since we shall elucidate, for instance, whether a program is the shortest one, using a time complexity bound $\mathsf{f}$. 
%
%
%
%
%
%
The following section takes a different approach that enables curriculum teaching effectively.

\section{Minimal curriculum: Interposition range and $\mathbb{I}$-search}\label{section5}


One key reason why interposition is hard to avoid is the existence of programs (and concepts) with {\em parallel behaviour}, i.e., programs with equal inputs-outputs up to large sizes of the inputs, e.g., one implementing the {\sf even} function, and the other doing the same except for the input $2^{300}$. However, in practice, the concepts we use in the break-out for a curriculum do not have this problem. 
For instance, we can use addition to teach multiplication. They coincide in a few cases, $2+2=4$ and $2\times2=4$, but they clearly differ in many other short inputs. 

Thus, let $a, b$ be distinct concepts such that $\exists (w_a, p_a), (w_b, p_b) \in \mathsf{f}-$Teaching Book, with $p_a$, $p_b$ in $L$ verifying $w_a \nvDash_\mathsf{f} p_b$ and $w_b \nvDash_\mathsf{f} p_a$. Assume that we use $w_a$ first and the learner outputs $p_a$, and adds it to $B={\{p_a\}}$. With this increased $L_B$, if we give $w_b$ to the learner, it does not output $p_a$ since $p_a \nvDash_{\mathsf{f}} w_b$.
However, there might still be interposition. For instance, suppose that $L$ has four instructions: $\mathsf{x}$, $\mathsf{y}$, $\mathsf{z}$ and $\mathsf{t}$. Let $B=\{\mathsf{xx}\}$ and suppose that $p_b=\mathsf{zytxz}$ is ${\mathsf{f}}$-compatible with $w_b$. Suppose that there exists $p=\mathsf{xxytxx}$, expressed as $p=@\mathsf{yt}@$ in $L_B$, such that $p \vDash_{\mathsf{f}} w_b$. Program $p$ would interpose to $p_b$. It would be important to know about such programs $p$, i.e., the ones that precede $p_b$ in $L_B$ and are posterior in $L$. 

\subsection{Interposition range: $\mathbb{I}$-sets}
%
%
%
Firstly, we define the set of {\em interposed programs}.

\begin{definition}
{\em Let $w$ be a witness set and $B$ be a library. Let $p$ be a program in language $L_B$ such that $p \vDash_{\mathsf{f}} w$. We define the $\mathbb{I}$-set of {\em interposed programs} in language $L_B$ for $p$ and 
$w$ as 
$\mathbb{I}_{w}^{\mathsf{f}}(p|B)=\{ q \text{ in } L_B : {q \vDash_\mathsf{f} w} \text{ and } q \prec p \}$}.
\end{definition}


We now show how large the $\mathbb{I}$-sets can be. To do that, we use the size of a program when its library calls are {\em unfolded}, i.e., given a program $p$ and a library $B$, we use $\circ(p)$ to denote the program that is equivalent to $p$ (as it worked in $L_B$) 
, where each primitive call $@$ has been replaced by the instructions of the called primitive in $B$.



Given an $\mathbb{I}$-set, we call {\em size-range}, denoted as $[i_{min}, i_{max}]$, to the range of $i=\ninst{(\circ(q))}$, $\forall q \in \mathbb{I}$-set. The {\em call-range}, denoted as $[j_{min}, j_{max}]$, is the range of the number of library calls, $j$, $\forall q \in \mathbb{I}$-set. We call {\em s/c-ranges} to both ranges; interposition occurs within them. The following theorem gives  the {\em s/c-ranges} explicitly and provides a bound for the cardinality of the $\mathbb{I}$-set.

\begin{theorem}\label{th_@}
{\em Let $(w_a, p_a)$, $(w_b, p_b) \in \mathsf{f}$-Teaching Book, with $p_a$, $p_b$ in $L$ and $p_a \nvDash_\mathsf{f} w_b$. Consider the library $B={\{p_a\}}$. Let $p_b'$ an equivalent program to $p_b$ for $L_B$. Then, the cardinal of $\mathbb{I}_{w_b}^{\mathsf{f}}(p_b'|B)$ is bounded by $\sum_i \big( \sum_j \binom{i-\ninst(p_b) \cdot j + j}{j} \cdot (|\Upsilon|-1)^{(i - j \cdot \ninst(p_b))} \big)$
with $i$, $j \in \mathbb{N}$ ranging in the intervals: (1) $i_{min}=\ninst(p_b)$, $i_{max}=1 + (\ninst(p_b')-1) \cdot \ninst(p_a)$, $j_{min}=\lceil{\frac{i-\ninst(p_b')}{\ninst(p_a)-1}\rceil}$ and $j_{max}= \lfloor{\frac{i}{\ninst(p_a)}\rfloor}$, when $1 < \ninst(p_a) < \ninst(p_b)$; (2) $i_{min}=\ninst(p_a)+1$ and the rest is as (1), when $\ninst(p_a) \geq \ninst(p_b)$.
}
\begin{proof}
\textbf{1st case:} We consider that the library does not reduce $p_b$, i.e., $p_b'=p_b$. In this 1st case, we prove that the cardinal of $\mathbb{I}_{w_b}^{\mathsf{f}}(p_b|B)$ is bounded by \newline
\begin{equation*}
    (|\Upsilon|-1) + 
    \sum_i \bigg( \sum_j \binom{i-\ninst(p_b) \cdot j + j}{j} \cdot (|\Upsilon|-1)^{(i - j \cdot \ninst(p_b))} \bigg),
\end{equation*}
where $i \in \bigg[ \ninst(p_b) , (\ninst(p_b)-1) \cdot \ninst(p_a) \bigg]$ and $j \in \Bigg[ \biggl\lceil{\frac{i-\ninst(p_b)}{\ninst(p_a)-1}\biggr\rceil} , \biggl\lfloor{\frac{i}{\ninst(p_b)}\biggr\rfloor} \Bigg]$.

We are interested in programs $q$ in $L_B$, which do not precede $p_b$ in $L$ but they do in $L_B$, i.e.,
\begin{enumerate}
    \item\label{L(q)_1} $\ninst(\circ(q)) \geq \ninst(p_b)$ (otherwise, $q \prec p_b$ in language $L$ and $q \nvDash_{\mathsf{f}} w_b$).
    \item\label{L(q)_2} $\ninst(\circ(q)) \leq (\ninst(p_b)-1) \cdot \ninst(p_a) + 1$ (otherwise $p_b \prec q$ in $L$).
\end{enumerate} 
    
The last condition \ref{L(q)_2} is equivalent to {\em fill} all but one instruction of $p_b$ with calls to the library. We do not consider the program $@@\cdots@$ of length $\ninst(p_b)$, since it is posterior or equal to $p_b$ ($@$ is the last instruction in lexicographical order).

Thus, $\forall q \in \mathbb{I}_{w_b}^\mathsf{f}(p_b|B)$ then
\begin{equation*}
    \ninst(p_b) \leq \ninst(\circ(q)) \leq (\ninst(p_b)-1) \cdot \ninst(p_a) + 1
\end{equation*}

Now, for each $i=\ninst(\circ(q))$, we study the number of allowed library calls, $j$, i.e.:  
\begin{enumerate}[resume]
    \item\label{L(q)_3} $j \leq \Bigl\lfloor{\frac{i}{\ninst(p_a)}\Bigr\rfloor}$. At most, there can be $j =i/\ninst(p_a)$ library calls, which is the case $i=j \cdot \ninst(p_a)$
    \item\label{L(q)_4} $j \geq  \Bigl\lceil{\frac{i-\ninst(p_b)}{\ninst(p_a)-1}\Bigr\rceil}$
\end{enumerate}

The last condition \ref{L(q)_4} occurs because any program in $L_B$, with higher priority than $p_b$, satisfies
\begin{equation*}
    \ninst(p_b) \geq (i-\ninst(p_a) \cdot j) + j
\end{equation*}
Since there are $j$ library calls and $i-\ninst(p_a) \cdot j$ instructions that do use $@$. 
In other words, condition \ref{L(q)_4} guarantees 
\begin{equation*}
   j \cdot (1-\ninst(p_a)) \leq \ninst(p_b) - i,
\end{equation*}
which implies:
\begin{equation*}
    j \geq \frac{\ninst(p_b) - i}{1-\ninst(p_a)}
\end{equation*}

That said, for each $q \in \mathbb{I}_{w_b}^\mathsf{f}(p_b|B)$, with $i=\ninst(\circ(q))$ and $j$ library calls, we have different distributions of the calls. Thus, for instance, programs like $@@\mathsf{xyz}$, $\mathsf{x}@\mathsf{y}@\mathsf{z}$ or $\mathsf{xy}@@\mathsf{z}$, employ the same number of instructions.

Once we fix $i=\ninst(\circ(q))$, we have to choose $j$ positions out of $(i-j \cdot \ninst(p_a)) + j$, i.e.,
\begin{equation*}
    \binom{(i - j \cdot \ninst(p_a)) + j }{j} \text{ or } \binom{(i - j \cdot \ninst(p_a)) + j }{i - j \cdot \ninst(p_a)}
\end{equation*}

Furthermore, for each one of these cases, we have $|\Upsilon|-1$ possible instructions for each one of the $(i - j \cdot \ninst(p_a))$ positions.

Therefore, given $i=\ninst(\circ(q))$, we get for each $j$ that verifies
\begin{equation*}
    \Biggl\lceil{\frac{i-\ninst(p_b)}{\ninst(p_a)-1}\Biggr\rceil} \leq j \leq \Biggl\lfloor{\frac{i}{\ninst(p_a)}\Biggr\rfloor}
\end{equation*}
a number of interposed programs less than
\begin{equation*}
    \sum_j \binom{(i - j \cdot \ninst(p_a)) + j}{j} \cdot (|\Upsilon|-1)^{(i-j \cdot \ninst(p_a))}
\end{equation*}

Even, the case $i=(\ninst(p_b)-1) \cdot \ninst(p_a)+1$ can be isolated, since interposed programs $q$ starting with $@$ in $L_B$ are lexicographically posterior to $p_b$ in $L_B$. In such a case, there are $|\Upsilon|-1$ programs. Finally, we get the conclusion of the 1st case.

\textbf{2nd case: } We prove part (1) of the theorem with $p_b'$ in $L_B$.

Now, the case $\ninst(\circ(q))=(\ninst({p_b}')-1) \cdot \ninst(p_a) + 1$, can not be isolated. Because ${p_b}'$ might begin with library calls. 

The rest is analogous to the 1st case of this proof.

\textbf{3rd case: } We prove part (2) of the theorem with $p_b'$ in $L_B$.

The proof is also analogous to the 1st case, but the only difference being is $i_{min}=\ninst(p_a)+1$. This is because a unique call for the library, $@$, is not compatible with $w_b$ and, with the translation to $L_B$, any program $q \in \mathbb{I}_{w_b}^\mathsf{f}({p_b}'|B)$ must call the library, so that $\ninst(\circ(q)) \geq \ninst(p_a)+1$.  
\end{proof}
\end{theorem}

%
{\em Could we identify an empty $\mathbb{I}$-set, based just on the sizes of the programs involved?} It happens when the 
{\em s/c-ranges} define an {\em empty region}. In 
Theorem \ref{th_@} (1), it occurs whenever $i_{max} < i_{min}$. Namely, we have $\mathbb{I}_{w_b}^\mathsf{f}({p_b}'|B) = \emptyset$, 
when:
\begin{equation}\label{condEmptyInterposition}
    \ninst(p_b) > 1 + (\ninst(p_b')-1) \cdot \ninst(p_a)
\end{equation}
For instance, if $\ninst(p_a)=4$, $\ninst(p_b)=8$ and we know that
$\ninst({p_b}')=2$, then $i_{min}=8$ and  $i_{max} = 1 + (2-1) \cdot 4=5$. We see that this becomes more likely as $p_b$ is much greater than $p_a$ and the program for $b$ using $B$, i.e., ${p_b}'$, is significantly reduced by the use of $B=\{p_a\}$.

Let $p'$ be 
the first program in $[b]_{L_B}$ such that $p' \vDash_{\mathsf{f}} w_b$. With the conditions of Theorem \ref{th_@} (1), $p'$ must be equivalent to $p_b$ and operating with Eq. \ref{condEmptyInterposition} we get
 $ \ninst(p') < \frac{\ninst(p_b) - 1}{\ninst(p_a)} +1$, which means there is no interposition for any program for $b$ by including $B=\{a\}$ and $TS_{\ell}^\mathsf{f}(b|a) \leq TS_{\ell}^\mathsf{f}(b)$.
But, since $\ell(p'_b) \geq K(b|a)$ we also have 
that  Eq. \ref{condEmptyInterposition} is impossible when 
$K(b|a) \geq (\frac{\ninst(p_b) - 1}{\ninst(p_a)} + 1)  \cdot \vepsilon$.



We now consider a library with more than one primitive. We cannot extend Theorem \ref{th_@} as a Corollary, since the relationships involved change completely, but we can connect both cases through the {\em s/c-ranges}. 

\begin{theorem}\label{th_@@}\sidenoteJ{Quedaría más claro si a $w_c$ aquí lo llamaras $w_b$ como en el anterior. Me ha confundido todo el rato, hasta hoy que ya lo veo todo claro.}
{\em Let  $\{(w_m, p_m)\}_{m=1}^{n}$, $(w_c, p_c)$ $\in \mathsf{f}$-Teaching Book, with $p_c$, $p_m$ in $L$, $\forall m$. 
Consider $B=\{p_m\}_{m=1}^{n}$ with $p_m \nvDash_{\mathsf{f}} w_c$, $\forall m$, and $1<|B|$.
\sidenoteJ{Me parece muy fuerte está cota, pero si dices que se necesaria.}
Let ${p_c}'$ be an equivalent program to $p_c$ for ${L_B}$. 
Let $D$, $r \in \mathbb{N}$ such that $\ell(p_c')=D \cdot \ell(\mathsf{@i}) + r$, i.e., they are the {\em divisor} and the {\em remainder} of the division $\ell(p_c')/\ell(\mathsf{@i})$. Note that $\ell(\mathsf{@i}) = \log_2|\Upsilon|  + \log_2|B|$. Let $p_{max}={\max}^{\prec} \{p_m\}_{1}^{n}$ and $p_{min}={\min}^{\prec} \{p_m\}_{1}^{n}$. Then, the cardinal of $\mathbb{I}_{w_c}^{\mathsf{f}}({p_c}'|B)$ is bounded by $|B| \cdot \sum_{s=2}^{\ninst(p_c')} \big( \sum_{t=1}^{s}  (|\Upsilon|-1)^{s-t} \cdot |B|^{t-1} \big)$ and the {\em s/c-intervals} are: (1) if $1 < \ninst(p_{min}) \leq \ninst(p_c)$, then $i_{min}=\ninst(p_c)$, $i_{max}=D \cdot \ninst(p_{max}) + \lfloor r/\vepsilon \rfloor$, $j_{min}= 
\lfloor \frac{\ninst(p_c') - \ninst(\circ(q))}{\ninst(\mathsf{@i})-\ninst(p_{max})} \rfloor$
and $j_{max}=\min \{ D, \lfloor \frac{\ninst(\circ(q))}{\ninst(p_{min})} \rfloor \}$; (2) if $\ninst(p_c)<\ninst(p_{min})$, then $i_{min}=\ninst(p_{min})+1$ and the rest is as in  (1). 

}

\begin{proof}
Firstly, we want to find an upper bound for $\mathbb{I}_{w_c}^\mathsf{f}({p_c}'|B)$. Programs without library calls are not interposed to ${p_c}'$, otherwise the non-incremental learner outputs a program different from $p_c$, which is a contradiction; we can only get interposition through library calls. 

Since interposed programs 
must employ equal or less instructions in language $L_B$ than ${p_c}'$, then $2 \leq \ninst(q) \leq \ninst({p_c}')$. Therefore, $|\mathbb{I}_{w_c}^\mathsf{f}({p_c}'|B)|$ is bounded by\newline 
\begin{equation*}
    \sum_{s=2}^{\ninst({p_c}')} (|\Upsilon|-1) + |B|)^s - \sum_{s=2}^{\ninst({p_c}')} (|\Upsilon|-1)^s =\sum_{s=2}^{\ninst({p_c}')} ((|\Upsilon|-1) + |B|)^s - (|\Upsilon|-1)^s
\end{equation*}

If we apply the binomial theorem and use $|B|$ as common factor, then we get the conclusion
\begin{equation*}
     |\mathbb{I}_{w_c}^\mathsf{f}({p_c}'|B)| \leq |B| \cdot \sum_{s=2}^{\ninst({p_c}')} \Big( \sum_{t=1}^{s}  (|\Upsilon|-1)^{s-t} \cdot |B|^{t-1}  \Big)    
\end{equation*}

\textbf{1st case: } We now obtain the {\em s/c-ranges} for part (1) of the theorem.

Interposed programs must satisfy $\ninst(\circ(q)) \geq \ninst(p_c)$, for each $q \in \mathbb{I}_{w_c}^\mathsf{f}({p_b}'|B)$. 

Now, let us show the maximum value of $\ninst(\circ(q))$. Every $q$ shall use the library and its priority must be higher than ${p_c}'$, i.e., $\ell(q) \leq \ell({p_c}')$.

The higher the number of library calls is, $j$, the larger $\ninst(\circ(q))$ is. We have, at most, $\ell({p_c}')$ bits, 
then the maximum number of library calls is $D$, where
\begin{equation*}
     \ell({p_c}') = D \cdot \ell(\mathsf{@i}) + r
\end{equation*}

The remainder $r$ can be used without library calls $@$, namely $\lfloor{r/\vepsilon}\rfloor$ instructions. Therefore:
\begin{equation*}
    \ninst(\circ(q)) \leq D \cdot \ninst(p_{max}) + \lfloor{r/\vepsilon}\rfloor
\end{equation*}

Now, once $\ninst(\circ(q))$ is fixed, we look for a lower bound of library calls, $j$. We know that, given $m \in \{1, \ldots, n\}$, then
\begin{equation}\label{j_min_inst}
    \ninst({p_c}') \geq j + (\ninst(\circ(q)) - \ninst(p_m) \cdot j)
\end{equation}

So that, since $\ninst(p_m)>1$, we get
\begin{equation}\label{ineq_j_min_inst}
     \biggl\lfloor{\frac{\ninst({p_c}') - \ninst(\circ(q))}{\ninst(@\mathsf{i}) - \ninst(p_{max})}\biggr\rfloor} 
     \leq
     \biggl\lfloor{\frac{\ninst({p_c}') - \ninst(\circ(q))}{1 - \ninst(p_m)}\biggr\rfloor} 
     \leq j
\end{equation}
Note that we take the higher denominator for all $p_m$, through $p_{max}$, to cover  every possibility. 

If we want to express inequality \ref{j_min_inst} in bits, it is
\begin{equation*}
    \ell({p_c}') \geq j \cdot \ell(\mathsf{@i}) + (\ninst(\circ(q)) \cdot \vepsilon - j \cdot \ninst(p_m) \cdot \vepsilon)
\end{equation*}

So that
\begin{equation*}
    \ell({p_c}') - \ninst(\circ(q)) \cdot \vepsilon \geq j \cdot (\ell(\mathsf{@i}) - \ninst(p_m) \cdot \vepsilon)
\end{equation*}

Suppose that

\begin{equation}\label{libraryrestrictions}
    \log_2|B| < \vepsilon \cdot( \ninst{(p_{max})}-1)
\end{equation}

Then $\ell(\mathsf{@i}) < \ninst(p_{max}) \cdot \vepsilon$
, we get another lower bound for library calls
\begin{equation}\label{j_min_bits}
     \biggl\lfloor{\frac{\ell({p_c}') - \ninst(\circ(q)) \cdot \vepsilon}{\ell(@\mathsf{i}) - \ninst(p_{max}) \cdot \vepsilon }\biggr\rfloor} \leq j
\end{equation}

We now consider the following expressions \ref{less_j}, \ref{bits_frac} and \ref{inst_frac}:
\begin{equation}\label{less_j}
    \frac{\ell(\circ(q)) - \ell(p_c')}{\ell(p_{max}) - \ell(\mathsf{@i})} \leq \frac{j \cdot \log_{2}{|B|}}{\log_{2}{|B|}}=j
\end{equation}
\begin{equation}\label{bits_frac}
    \frac{\ell({p_c}') - \ninst(\circ(q)) \cdot \vepsilon}{\ell(@\mathsf{i}) - \ninst(p_{max}) \cdot \vepsilon }
    =
    \frac{\ell(\circ(q)) - \ell(p_c')}{\ell(p_{max}) - \ell(\mathsf{@i})} 
\end{equation}   
\begin{equation}\label{inst_frac}
   {\frac{ \ell(\circ(q)) - \ell(p_c') +  j \cdot \log_{2}{|B|} }{\ell(p_{max}) - \ell(\mathsf{@i}) + \log_{2}{|B|}}} = \frac{\ninst({p_c}') - \ninst(\circ(q))}{\ninst(@\mathsf{i}) - \ninst(p_{max})}
\end{equation}  

Also, we consider the following property:
\begin{equation}\label{frac_prop}
  \forall \frac{x}{y}, \frac{u}{v} \text{ such that }\frac{x}{y} \leq \frac{u}{v} \text{ then } \frac{x}{y} \leq \frac{x+u}{y+v} 
\end{equation}

If we consider the inequality \ref{less_j} and property \ref{frac_prop}, then we can {\em connect} Eq. \ref{bits_frac} and Eq. \ref{inst_frac}, through the following inequality: 
\begin{equation}
\frac{\ell(\circ(q)) - \ell(p_c')}{\ell(p_{max}) - \ell(\mathsf{@i})} 
\leq 
{\frac{ \ell(\circ(q)) - \ell(p_c') +  j \cdot \log_{2}{|B|} }{\ell(p_{max}) - \ell(\mathsf{@i}) + \log_{2}{|B|}}}
\end{equation}

Therefore, inequality \ref{ineq_j_min_inst} improves \ref{j_min_bits} as a lower bound:
\begin{equation*}
     \biggl\lfloor{\frac{\ell({p_c}') - \ninst(\circ(q)) \cdot \vepsilon}{\ell(@\mathsf{i}) - \ninst(p_{max}) \cdot \vepsilon }\biggr\rfloor} 
     \leq 
      \biggl\lfloor{\frac{\ninst({p_c}') - \ninst(\circ(q))}{\ninst(\mathsf{@i}) - \ninst(p_{max})}\biggr\rfloor} 
     \leq j
\end{equation*}

If we use inequality \ref{ineq_j_min_inst} as a lower bound, there is no restriction for the number of primitives in the library. The assumptions (\ref{libraryrestrictions}) for the library are just to assure inequality \ref{j_min_bits}.

Now we look for an upper bound of library calls $j$, given $\ninst(\circ(q))$. On one side, there is a limit for the number of instructions in $L$ 
\begin{equation*}
    j \leq \biggl\lfloor{\frac{\ninst(\circ(q))}{\ninst(p_{min})}\biggr\rfloor}
\end{equation*}

But, the equivalent limit expressed in bits is $D$, i.e., 
\begin{equation*}
    j \leq \biggl\lfloor{\frac{\ell({p_c}')}{\ell(\mathsf{@i}))}\biggr\rfloor} = D
\end{equation*}

Since, every $q$ shall meet both limits, then 
\begin{equation*}
    j \leq min \Bigg\{ D, \biggl\lfloor{\frac{\ninst(\circ(q))}{\ninst(p_{min})}\biggr\rfloor} \Bigg\} 
\end{equation*}

\textbf{2nd case: } We now obtain the {\em s/c-ranges} for part (2) of the theorem.

We know that, each $q \in \mathbb{I}_{w_c}^\mathsf{f}({p_c}'|B)$ shall be of size greater than $p_c$ in $L$. In other words, $\ninst(\circ(q)) \geq \ninst(p_c)$, otherwise the learner outputs $q$ on input $w_c$, which is a contradiction. 

The main difference with part (1) is that $\ninst({p_c}') \leq \ninst(p_c)$, so that, any interposed program $q$ in $L_B$, which does not employ the library, cannot cause interposition. Only programs with instructions $\mathsf{@i}$ can be interposed. Since $\ninst(p_{min})>\ninst(p_c)$, then $\circ(\mathsf{@i})>\circ(p_c)$. So that, just a call to the library would cause interposition, had it not been because  $\mathsf{@i} \nvDash_{\mathsf{f}} w_c$. That is why an interposed program needs, at least, two instructions in language $L_B$. Therefore, $\ninst(p_{min})+1 \leq \ninst(\circ(q))$.

The rest of the proof is similar to the 1st case.
\end{proof}
\end{theorem}


We need $D \cdot \ninst(p_{max}) + \lfloor{r/\vepsilon}\rfloor < \ninst(p_{min}) + 1$, to avoid interposition directly, in the same conditions as in Theorem \ref{th_@@} (1). 
It entails $\ell({p_c}')<\ell(\mathsf{@i})$ when $\lfloor{r/\vepsilon}\rfloor = 0$ in the extreme case.   
For Theorem \ref{th_@@} (2), an unfeasible {\em s-range} implies $D< \frac{\ninst(p_c) - \lfloor{r/\vepsilon}\rfloor}{\ninst(p_{max})}$, which is  restrictive.




\subsection{Teaching size upper bounds: $\mathbb{I}$-safe}

In practice, we deal with a program $p$ that 
has the desired behaviour for a given witness set, but there may be interposition. If we know which the interposed programs are, then it is possible to 
get an upper bound of the teaching size of the concept that defines $p$, by {\em deflecting} interposition, refining the witness sets. 

We employ {\em $\mathbb{I}$-safe} witnesses: example sets attached to input/output pairs. For instance, if we want to teach exponentiation, a set of examples might be $\{(3,1) \rightarrow 3, (2,2) \rightarrow 4\}$. This witness set is compatible with exponentiation, but also compatible with multiplication. To avoid multiplication being interposed, we can add another example to distinguish both concepts: $\{(3,1)\rightarrow 3, (2,2)\rightarrow 4, (2,3)\rightarrow 8\}
$. 
We can always replace the original witness set by an $\mathbb{I}$-safe witness set, where, in general, we need to add examples to avoid interposition. 







\begin{proposition}\label{i-safe}
{\em Let $\mathsf{f}$ be a complexity function and $(w, p)$, $\{(w_m, p_m)\}_{m=1}^{n}$ $\in \mathsf{f}$-Teaching Book, with $p$, $p_m$ in $L$, $\forall m$. Let $B=\{p_m\}_{m=1}^{n}$ be a library such that $p_m \nvDash_{\mathsf{f}}
w$, $\forall m$. Let $c \in C_L$ such that $c \vDash w$. Let $p_c' \in [c]_{L_B}$ be the first program, using order $\prec$, such that $p_c' \vDash_{\mathsf{f}} w$. If $n=|\mathbb{I}_{w}^{\mathsf{f}}(p_c'|B)|$
, there exist $\{\langle \mathtt{i}_k,\mathtt{o}_k\rangle\}_{k=1}^{n}$ 
such that $TS_{\ell}^{\mathsf{f}}(c|B)\leq \delta\bigl(w \bigcup_{k=1}^{n} \{\langle \mathtt{i}_k,\mathtt{o}_k\rangle\}\bigr)$.}
\begin{proof}
Let us enumerate the interposed programs: $\mathbb{I}_{w}^\mathsf{f}(p_c'|B) = \{q_k\}_{k=1}^n$, such that $q_k \prec q_{k+1}, \forall k$.


For each $k$, we define what we call an $\mathbb{I}$-safe: the first input-output pair using $\lessdot$, $w_k=\{\langle \mathtt{{i}}_k, \mathtt{{o}}_k \rangle\}$, such that

\begin{equation*}
    p_c' \vDash_\mathsf{f} w_k \text{ and } q_j \nvDash_\mathsf{f} w_j
\end{equation*}

Note that, we define $\mathtt{{o}}_k=p_c'(\mathtt{{i}}_k)$ after generating $\mathtt{{i}}_k$ through $\lessdot$.

Finally, we aggregate every $\mathbb{I}$-safe pair to $w$. 
\end{proof}
\end{proposition}

For a library $B$, if we find 
an example set $w$ that can be converted into an $\mathbb{I}$-safe witness set $\overline{w}= w \bigcup_{k=1}^{n} \{\langle \mathtt{i}_k,\mathtt{o}_k\rangle\}$ with $\delta(\overline{w}) < TS^{\mathsf{f}}_{\ell}(c)$ using $B$, 
then we 
reduce 
the teaching size. This is a sufficient and necessary condition to avoid interposition and get $TS^{\mathsf{f}}_{\ell}(c|B) \leq TS^{\mathsf{f}}_{\ell}(c)$.  
%
%

Finally, given these general bounds: {\em how can we find minimal curricula?} Let us consider, for example, the set of concepts $Q=\{a, b\}$, where $(w_a, p_a)$ and $(w_b, p_b)$ are in the $\mathsf{f}$-Teaching Book. We also know that their behaviours are not parallel, i.e., $p_a \nvDash_\mathsf{f} w_b$ and $p_b \nvDash_{\mathsf{f}} w_a$. There are three different curricula $\{ a , b \}$, $\{ a \rightarrow b \}$ or $\{ b \rightarrow a \}$. 
There is an $\mathbb{I}$-safe witness set $\overline{w}$, such that $\delta(\overline{w}) \leq TS^{\mathsf{f}}_{\ell}(b|a)$ (or $\delta(\overline{w}) \leq TS^{\mathsf{f}}_{\ell}(a|b)$). Thus, we can choose a curriculum, with less overall teaching size than the non-incremental version.

\subsection{Minimal curriculum algorithm: $\mathbb{I}$-search}

We now {\em search} minimal curricula. For example, let $Q=\{c_{+}, c_{\times}\}$ be a set of two concepts from Fig. \ref{fig:curriculum}, which appear in the non-incremental $\mathsf{f}$-Teaching Book as $(w_{+}, p_{+})$ and $(w_{\times}, p_{\times})$. The set of possible curricula, $\overline{Q}$, is $\pi_0=\{c_{+}, c_{\times}\}$, $\pi_1=\{c_{+} \rightarrow c_{\times}\}$ and $\pi_2=\{c_{\times} \rightarrow c_{+}\}$. 

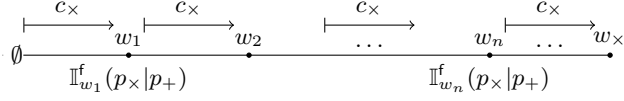
\begin{figure}
    \centering
    \begin{tikzpicture}[scale=0.8]
\draw (-0.75,0) -- (1,0);
\draw (1,0) -- (3, 0);
\draw (3, 0) -- node[above]{$\ldots$} (7,0);
\draw (7,0) -- node[above]{$\ldots$} (9,0);

\draw[|->](-0.75,0.5) -- node[above]{$c_{\times}$}(0.75,0.5);
\filldraw[black] (-1.1,0) circle (0pt) node[anchor=west] {$\emptyset$};
\filldraw[black] (1,0) circle (1pt) node[anchor=south] {$w_1$};
\filldraw[black] (1,0) circle (0pt) node[anchor=north] {$\mathbb{I}_{w_1}^{\mathsf{f}} (p_{\times}|p_{+})$};
\draw[|->](1.25,0.5) -- node[above]{$c_{\times}$}(2.75,0.5);
\filldraw[black] (3,0) circle (1pt) node[anchor=south] {$w_2$};
\filldraw[black] (3,0) circle (0pt) 
;
\draw[|->](4.25,0.5) -- node[above]{$c_{\times}$}(5.75,0.5);
\filldraw[black] (7,0) circle (1pt) node[anchor=south] {$w_n$};
\filldraw[black] (7,0) circle (0pt) node[anchor=north] {$\mathbb{I}_{w_n}^{\mathsf{f}} (p_{\times}|p_{+})$}
;

\draw[|->](7.25,0.5) -- node[above]{$c_{\times}$}(8.75,0.5);
\filldraw[black] (9,0) circle (1pt) node[anchor=south] {$w_{\times}$};
\filldraw[black] (9,0) circle (0pt) 
;
\end{tikzpicture}
\caption{Non-decreasing sequence of witness sets $w_k$, through $c_{\times}$ with $\delta(w_k) \leq \delta(w_{\times})$.}
    \label{fig:ts_w_i}
\end{figure}

The starting point for our algorithm will be $\pi_0$, the non-incremental curriculum, and its overall teaching size $TS_{\ell}^{\mathsf{f}}$. Then, we generate another curriculum: $\pi_1$. We know $TS_{\ell}^{\mathsf{f}}(c_{+})=\delta(w_{+})$ and we need to add  $TS_{\ell}^{\mathsf{f}}(c_{\times}|c_{+})$. We compare this total size to the best TS so far. We explore all the curricula in $\overline{Q}$ but, in order to save computational steps, we generate successive witness sets $w_k$, using order $\lessdot$, such that $c_{\times} \vDash w_k$ (Fig. \ref{fig:ts_w_i}). For each $w_k$, we get the first program $p_k$ of $\mathbb{I}_{w_k}^{\mathsf{f}}(p_{\times}| p_{+})$. We then investigate whether $p_k \in [p_{\times}]_{L_B}$ or not. If $p_k$ acts like $p_{\times}$ to certain witness size limit, $H$, then we can identify 
 $p_k$ and $p_{\times}$. 
The following algorithm extends this strategy in general: 

\noindent
\fbox{
\begin{minipage}{33.5em}
\begin{small}
  \textbf{Algorithm: } $\mathbb{I}$-search \newline
  \textbf{Input: } $Q=\{a, b, \ldots\}$; 
  $\mathsf{f}$-Teaching Book $(w_a, p_a)$, $(w_b, p_b)...$; Witness size limit $H$
  \vspace{-0.2cm}
  \begin{enumerate}
  \item\label{sieve} \textbf{For each} distinct pair of concepts $\langle x, y \rangle \in Q \times Q$\textbf{:}
        \begin{enumerate}
            \item \textbf{If [}$TS_{\ell}^{\mathsf{f}}(y|x) \leq TS_{\ell}^{\mathsf{f}}(y)$ $\wedge$ $TS(x|y)_{\ell}^{\mathsf{f}} \geq TS_{\ell}^{\mathsf{f}}(x)$\textbf{]}
            \\ 
            \textbf{then} 	$\overline{Q} = \overline{Q} \setminus \{\pi: \exists \text{ a branch starting as }y\rightarrow x\}$
        \end{enumerate}
  \item $\pi^*=\{a, b, \ldots\}$,  $TS_{\ell}^{\mathsf{f}}(\pi^*)= \sum_{x \in Q} TS_{\ell}^{\mathsf{f}}(x)$ and $\overline{Q} = \overline{Q} \setminus \{ \pi^* \}$
  \item\label{new_curriculum} \textbf{For each} $\pi \in \overline{Q}$\textbf{:}
    \begin{enumerate}
        \item $TS_{\ell}^{\mathsf{f}}(\pi)=0$
        \item \textbf{For each} branch $\sigma \in \pi$\textbf{:}
        \begin{enumerate}
            \item\label{new_concept} \textbf{For each} concept $x \in \sigma$ (ordered by $\sigma$)\textbf{:}
            \begin{itemize}
                \item $B=\{p_y: (y \in \sigma) \wedge (y \text{ precedes }
                x)\}$
                \item Let $p_x'$ be the first program equivalent to $p_x$ in $L_B$, using order $\prec$
                    \item \textbf{For each} $w_k \in \{w \subset X: p_x' \vDash_{\mathsf{f}} w_k\}$, using order $\lessdot$\textbf{:}
                    \begin{itemize}
                        \item \textbf{If [}$TS_{\ell}^{\mathsf{f}}(\pi^*) \leq TS_{\ell}^{\mathsf{f}}(\pi) + \delta(w_k) $\textbf{]} \textbf{then} \textbf{break} to \ref{new_curriculum} 
                        \item $p = \min^{\prec}\{\mathbb{I}_{w_k}^{\mathsf{f}}(p_x'|B)\}$; use {\em s/c ranges} to refine the calculation
                        \item \textbf{If  [}$p \vDash_{\mathsf{f}} w \longleftrightarrow p_x \vDash_{\mathsf{f}} w$, $\forall w$ such that 
                        $\delta(w)<H$\textbf{]} \\ \textbf{ then} \textbf{[} $TS_{\ell}^{\mathsf{f}}(\pi)=TS_{\ell}^{\mathsf{f}}(\pi) + \delta(w_k)$ \textbf{ and break} to \ref{new_concept} \textbf{]}
                    \end{itemize}
            \end{itemize}
        \end{enumerate}
        \item $\pi^*=\pi$ and $TS_{\ell}^{\mathsf{f}}(\pi^*)=TS_{\ell}^{\mathsf{f}}(\pi)$
    \end{enumerate}
  \end{enumerate}
  \vspace{-0.2cm}
  \textbf{Output: } $\pi^*$ and $TS_{\ell}^{\mathsf{f}}(\pi^*)$
\end{small}
\end{minipage}
}\label{I-search}

\sidenoteJ{Te he quitado la footnote en el algoritmo, era muy vaga: "might help". O ayuda o no ayuda, y tampoc explica cómo. En todo caso en el texto}
Note that the {\em s/c-ranges} reduce, {\em drastically}, the computational effort of executing the teacher-learner protocol (calculating teaching book and TS). 
In the previous example, e.g., if there is a $w_n$ such that $TS_{\ell}^{\mathsf{f}}(c_{\times}|c_{+})=\delta(w_n) < TS_{\ell}^{\mathsf{f}}(c_{\times})$, then we set $\pi^*=\pi_1$ (and $TS_{\ell}^{\mathsf{f}}(\pi^*) = \delta(w_{+})+\delta(w_n)$). 
Finally, we test $\pi_2$ and follow the same steps as with $\pi_1$. If, at some stage, there is a witness set $w_m$ such that $TS_{\ell}^{\mathsf{f}}(c_{\times})+\delta(w_m) \geq TS_{\ell}^{\mathsf{f}}(\pi^*)$, then 
$\pi_1$ is minimal and we stop.

The algorithm is complete but the search is not {\em exhaustive}, 
since we can {\em discard} curricula that contain a {\em branch} starting in a way that does not decrease the overall teaching size for sure. For example, if $TS_{\ell}^{\mathsf{f}}(c_{\times}|c_{+}) \leq TS_{\ell}^{\mathsf{f}}(c_{\times})$ and $TS_{\ell}^{\mathsf{f}}(c_{+}|c_{\times}) \geq TS_{\ell}^{\mathsf{f}}(c_{+})$, the branch $\sigma=\{c_{+} \rightarrow c_{\times} \rightarrow c_{\wedge}\}$ 
has less or equal 
overall teaching size than $\sigma'=\{c_{\times} \rightarrow c_{+} \rightarrow c_{\wedge} \}$. Consequently, we can remove all branches starting with $c_{\times} \rightarrow c_{+}$. We can test this for every pair of distinct concepts at the beginning of the branches. 

The {\em $\mathbb{I}$-search algorithm} (\ref{I-search}) satisfies the following theorem.

\begin{theorem}
{\em 
Let $H$ be certain witness size limit, $\mathsf{f}$ be a complexity function and $Q$ be a set of concepts registered in the $\mathsf{f}$-Teaching Book. We also assume, for each $c \in Q$, that $c \vDash w \rightarrow p_c \vDash_{\mathsf{f}} w$, $\forall w$ verifying $\delta(w) \leq \sum_{x \in Q}TS_{\ell}^{\mathsf{f}}(x)$\sidenoteM{Creo que esta condición es necesaria para garantizar que los programas equivalentes a un $p_c$ en cada $L_B$ tendrán el mismo comportamiento que $p_c$ en todo $w$ que no supere el tamaño del curriculum no incremental.}\sidenoteJ{Me fío de ti :-)}\sidenoteM{Es que con esto me lío. A ver si me aclaro: El programa $p_c$ que escoge el f-T.Book es compatible con todos los $w$ que pertenecen a un concepto $c$, pero ¿puede que haya algún $w'$ que, aunque $c \vDash w'$, pase $p_c \nvDash_{\mathsf{f}} w'$ por no llegar al límite de tiempo? En caso afirmativo, sí necesitamos esta propiedad.}\sidenoteJ{Me llevas un lío con w y w', y con $p_c \nvDash_{\mathsf{f} f}$ ???? }. Then, the $\mathbb{I}$-search algorithm expressed in algorithm  \ref{I-search} returns a minimal curriculum and its overall teaching size. 
}
\begin{proof}

Firstly, we want to discard some curricula.

Let $x$, $y \in Q$ such that $TS_{\ell}^{\mathsf{f}}(y|x) \leq TS_{\ell}^{\mathsf{f}}(y)$ and $TS_{\ell}^{\mathsf{f}}(x|y) \geq TS_{\ell}^{\mathsf{f}}(x)$ then:
\begin{equation}\label{x-y}
     TS_{\ell}^{\mathsf{f}}(x)+TS_{\ell}^{\mathsf{f}}(y|x) \leq  TS_{\ell}^{\mathsf{f}}(x)+TS_{\ell}^{\mathsf{f}}(y)
\end{equation}
and
\begin{equation}\label{y-x}
     TS_{\ell}^{\mathsf{f}}(y)+TS_{\ell}^{\mathsf{f}}(x|y) \geq  TS_{\ell}^{\mathsf{f}}(x)+TS_{\ell}^{\mathsf{f}}(y)
\end{equation}
We get $TS_{\ell}^{\mathsf{f}}(x)+TS_{\ell}^{\mathsf{f}}(y|x) \leq TS_{\ell}^{\mathsf{f}}(y)+TS_{\ell}^{\mathsf{f}}(x|y)$, using inequalities \ref{x-y} and \ref{y-x}. Therefore, a branch starting with $y \rightarrow x$ cannot {\em improve} another branch starting as $x \rightarrow y$. 

The order $\prec$ guarantees that the first index of the library points to its first program. As a result, the programs that the incremental learner builds, after teaching $x$ and $y$, does not change whether the teaching order is $x \rightarrow y$ or $y \rightarrow x$.

We can repeat this procedure for every two concepts in $Q$ and it might reduce the number of candidates to minimal curricula.

Secondly, we take $\pi_0=\{a, b, \ldots \}$, the non-incremental curriculum, as a reference and we set $\pi^*=\pi_0$ and $TS_{\ell}^{\mathsf{f}}(\pi^*)=\sum_{x \in \pi_0}TS_{\ell}^{\mathsf{f}}(x)$. We now take a different curriculum, $\pi_1 \in \overline{Q}$ and we check whether
\begin{equation}\label{improve_curriculum}
    TS_{\ell}^{\mathsf{f}}(\pi_1) < TS_{\ell}^{\mathsf{f}}(\pi^*)
\end{equation}
We do it by following the curriculum's branches and adding, successively, the teaching size of its concepts. If, at some stage of the process, inequality  \ref{improve_curriculum} is false, then $\pi_1$ cannot improve $\pi^*$ and we take another curriculum $\pi_2$. Otherwise, if inequality \ref{improve_curriculum} is true, then we set:
\begin{equation*}
    \pi^*=\pi_1 \text{ and } TS_{\ell}^{\mathsf{f}}(\pi^*)=TS_{\ell}^{\mathsf{f}}(\pi_1)
\end{equation*}

There are two important issues in this part of the algorithm:
\begin{itemize}
    \item Let $p_x'$ be the first program, using order $\prec$, which is equivalent to $p_x$ in $L$, at some stage of the procedure. For each branch, $\sigma \in \pi$, each $x \in \sigma$ and each $w_k \in \{w \subset X: p_x' \vDash_{\mathsf{f}} w_k\}$, we calculate the first program of $\{\mathbb{I}_{w_k}^{\mathsf{f}}(p_x'|B)\}$, using order $\lessdot$. 
    \begin{itemize}
        \item If $\mathbb{I}_{w_k}^{\mathsf{f}}(p_x'|B) = \emptyset$, then we 
        exit the loop we are inside and move to \ref{new_concept}.
        \item Otherwise, we get a program $p$ with higher priority than $p_x'$. 
        \item If the behaviour of $p$ and $p_x$ is equal until certain witness size limit, $H$, then we can identify both programs and move to  \ref{new_concept}. Otherwise, we get the next witness set.
    \end{itemize}
    \item We could have problems if there is a witness set $w_k$, such that $p_x' \nvDash_{\mathsf{f}} w_k$. We should look for another equivalent program, but it might be that $\mathsf{f}$ is not sufficient, i.e., we shall find another program $\mathsf{f}$-compatible with $w_k$. All in all, it would be quite similar to the teacher-learner protocol. That is why, in order to avoid this issue, we assumed that $p_c \vDash_{\mathsf{f}} w$, $\forall w$ with $c \vDash w$. We need this assumption only for witness sets $w$ such that $\delta(w) \leq \sum_{x \in Q}TS_{\ell}^{\mathsf{f}}(x)$, since we cannot get higher overall teaching size.
\end{itemize}

We proceed succesively with $\pi_2$, $\pi_3$ and so on. In the end, the algorithm returns a minimal curriculum and its overall teaching size.
\end{proof}
\end{theorem}

The $\mathbb{I}$-search algorithm shows that:
(1) We should create curricula containing concepts that significantly reduce the complexity of another ones. For instance, if concepts $c_{\times}$ and $c_{+}$ (Fig. \ref{fig:curriculum}) satisfy $K(c_{\times}|c_{+})<K(c_{\times})$, then the chances to minimise the teaching size increase significantly. 
(2) Given a set of concepts, it may be useful to implement some kind of {\em isolation} (or even forgetting by separating concepts in different branches\footnote{Forgetting may simply refer to a lesson not using primitives that are considered out of the context of a ``lesson''.}). For instance,  $c_{\stkout{0}}$ might be $\mathsf{f}$-compatible with a considerable number of witness sets $w_k$ and it may cause {\em interposition} with $c_{+}$, $c_{\times}$ or $c_{\wedge}$. This is why we should allocate $c_{\stkout{0}}$ in a different branch.
(3) The branches (or lessons) could simply suggest ways in which we can arrange, {\em classify} and organise large sets of concepts. The tree-structure for curricula proposed here is a solution for the problem posed in \cite{Pentina_2015_CVPR}.

\section{Conclusions and future work}\label{section6}

The teaching {\em size} ---rather than teaching dimension--- opened a new avenue for a more realistic and powerful analysis of machine teaching \cite{Telle2019TheTS}, its connections with information theory (both programs and examples can be measured in bits) and a proper handling of concept classes where examples and programs are  compositional and possibly universal, such as natural language.


The intuitive concept of how much of the description of a concept is reused for the definition of another dates back to Leibniz's {\em règle pour passer de pensée en pensée} \cite{leibniz2018}, 
and has been vindicated in cognitive science
since Vigotsky's zone of proximal development \cite{vigotsky1978,theorieshumandevelopment}, to more modern accounts of compositionality based on what has been learnt previously \cite{oberauer2017,manohar2019,schneider2020}.\sidenoteM{Evidence suggests both sustained activity and synaptic plasticity support working memory. (Manohar, 2019)}\sidenoteM{how cognitive behavior of humans, animals, and
machines with its key features of flexibility and context-sensitivity are realized at the functional and
mechanistic level}\sidenoteM{Objective: how cognitive behavior of humans, animals, and
machines with its key features of flexibility and context-sensitivity are realized at the functional and
mechanistic level(Scheneider, 2020)}\sidenoteM{The success of the interference model shows that working memory for continuous visual information works according to the same principles as working memory for more discrete (e.g., verbal) contents (Oberauer, 2017)}
%

In mathematical terms, a gradient-based or continuous account of this view of incremental teaching, and the reuse of concepts, is not well accommodated. Incremental teaching is usually characterised as a compositional process, which is a more appropriate view for the acquisition of high-level concepts. The learning counterpart is still very elegantly captured by conditional Kolmogorov complexity, and some incremental learning schemata have followed this inspiration \cite{lake2015human,gulwani2015inductive,shi2018concept,li2020perspective,nye2020learning}. 
However, even if the concept of teaching {\em size} suggests that a mapping was possible, we have had to face a series of phenomena in order to translate some of these intuitions to the machine teaching scenario, and a new setting for curriculum teaching. 

The absence of monotonicity because of interposition
presents some difficulties for implementing curriculum teaching for compositional languages. 
Theorems \ref{th_@} and \ref{th_@@} and its consequences make possible such an implementation: either through sufficient conditions to avoid interposition, 
by implementing $\mathbb{I}$-safe witness sets or through the $\mathbb{I}$-search. 
%
%
%

Given the theoretical bounds and the algorithms for the optimal curricula, we can now start exploring novel algorithms and strategies for curriculum teaching that are suboptimal, but more efficient, such as (1) greedy algorithms introducing the next concept as the one with maximum local TS reduction, (2) approximations based on Vigotsky's zone of proximal development principles \cite{vigotsky1978,theorieshumandevelopment} where each step is bounded by some teaching length $Z$, i.e., such that $TS(c_{i+1}|c_1, \ldots, c_{i}) \leq Z, \forall i $; or (3) variations of the \textit{incremental combinatorial optimal path} algorithm \cite{combinatorialoptimalpath}.
All these new research possibilities in curriculum teaching, and even others, are now wide open to exploration.

Because of the fundamental (re-)connection we have done between K and TS in this paper, another novel possibility for curriculum teaching would be the combination of teaching by examples {\em and} descriptions of the concepts themselves. This is actually the way humans teach other humans, combining examples and descriptions, but it is nevertheless unprecedented in the application of machine teaching in natural language processing  \cite{peng2020soloist,shukla2020conversation}. However, it is beginning to become common with language models, with prompts that combine examples and some indications of the task to perform \cite{brown2020language,hendrycks2020measuring}.

\section*{Acknowledgements}
This work was funded by the
EU (FEDER) and Spanish MINECO under RTI2018-094403-B-C32, G. Valenciana under PROMETEO/2019/098 and  
EU's Horizon 2020 research and innovation programme under grant 952215 (TAILOR). 

%
%
%
\bibliographystyle{splncs04}
\bibliography{mybibliography}
%




\end{document}